\ifdefined\pdfoutput\pdfoutput=1\fi
\documentclass[11pt]{article}

\usepackage[preprint]{acl}

\usepackage{times}
\usepackage{latexsym}
\usepackage[T1]{fontenc}
\usepackage[utf8]{inputenc}
\usepackage{microtype}
\usepackage{inconsolata}
\usepackage{graphicx}
\usepackage{booktabs}
\usepackage{amsmath}
\usepackage{amssymb}
\usepackage{xcolor}
\usepackage{multirow}


\graphicspath{{figures/}}

\newcommand{\basecond}{\textsc{base}}
\newcommand{\odonor}{\textsc{O-donor}}
\newcommand{\mapdonor}{\textsc{map-donor}}
\newcommand{\mapnull}{\textsc{map-null}}
\newcommand{\test}{\textsc{test}}
\newcommand{\route}{\textsc{route}}

\title{Test, then Route: How Language Models Execute In-Context\\
Conditional Rules Across Models and Languages}

\author{
  \textbf{Luxshan Thavarasa}\textsuperscript{1},\;
  \textbf{Sivasuthan Sukumar}\textsuperscript{2} \\[4pt]
  \textsuperscript{1}\,Independent Researcher, Colombo, Sri Lanka \\
  \textsuperscript{2}\,Department of Computer Science and Engineering, University of Moratuwa, Sri Lanka \\[4pt]
  {\small\texttt{luxshanlux2000@gmail.com}, \texttt{sivasuthan.26@cse.mrt.ac.lk}}}

\begin{document}
\maketitle

\begin{abstract}
When a language model follows an in-context conditional rule such as ``if $P(x)$ then $A$ else $B$,'' does it assemble a runtime circuit with one module that \emph{tests} the predicate and another that \emph{routes} the answer? We probe this with activation patching under a four-donor design whose two swapped-rule donors make the condition and the answer word \emph{disagree}, so each layer reveals which of the two it carries. Across three open models from two families and six languages sharing one fixed item bank, a mid-stack residual band carries the predicate's truth value: patching it reroutes the answer with predicate-outcome flip near $1.0$ and mapping flip near $0.0$, meeting a strict pre-specified isolation criterion in 17 of 18 cells, and the same localization holds across five predicate families. The router shows the opposite profile. A learned subspace flips $A{\leftrightarrow}B$ near-perfectly within the trained pair yet transfers to a new pair at ${\approx}0$ in every model, while in Gemma-3-4B (the only model probed cross-lingually) it transfers at ${\approx}0.98$ to the \emph{same} pair in other languages. Under every probe we ran, the router direction is token-bound and non-transferable (largely answer-readout in Gemma, pair-specific in Qwen) rather than an abstract routing module. Test is modular; under these probes, route is not.

\end{abstract}

\section{Introduction}

A large language model given an in-context conditional rule (``if $P(x)$ then $A$ else $B$'') must do two things to answer a query: \emph{test} whether the predicate $P(x)$ holds, and \emph{route} the appropriate answer, $A$ or $B$, to the output. Whether the model assembles these as separable parts, or entangles testing and routing in one indivisible mechanism, is a basic question about how in-context rules are executed, and one that behavioral evaluation cannot answer: a correct completion is consistent with either organization.

Most mechanistic circuits characterized to date are \emph{weight-stored}: pretrained, reusable mechanisms invoked for a fixed task, such as the Indirect Object Identification circuit \citep{ioi_circuit} or induction heads \citep{induction_heads}. The conditional rule is different: its text, predicate, and answer mapping are all supplied at inference time and vary from prompt to prompt, so any circuit executing it must be \emph{assembled at runtime}. This places our object of study alongside function and task vectors \citep{function_vectors,task_vectors} and filter heads \citep{filter_heads_lists}, but shifts the question from \emph{what} in-context content is represented to \emph{how} a two-part conditional computation is factored across the residual stream. Deployed prompts are dense with such conditionals (guardrails, tool-routing, conditional policies; Appendix~\ref{app:realistic}), and which half is modular determines what can be audited and steered portably.

We study a canonical rule (``if the number is greater than 5, output dog, otherwise cat''), presented with eight demonstrations and a numeric query whose single-token answer the model completes. Our central move is a set of four donor prompts (Figure~\ref{fig:donors}) that, through activation patching, read off \emph{per layer} whether a residual-stream site carries the predicate's truth \emph{condition} or the routed \emph{answer} word. The two critical donors make the condition and the answer \emph{disagree}: the \mapdonor{} swaps the rule's answer mapping while holding the predicate true; the \mapnull{} does so while holding it false. A layer that follows the truth value holds the condition; one that follows the answer word holds the readout. This exposes a sharp stage boundary (Figure~\ref{fig:stages}): a mid-stack \test{} stage where patches follow the condition, and a later \route{} stage where they follow the answer.

Applying this, we find a clear asymmetry. The predicate-\test{} is a causally separable, localizable module: patching a single mid-stack band flips the answer by flipping only the predicate's truth value. Under every probe we ran, the argument-\route{} shows no separable, transferable subspace: the direction that appears to route the answer flips one label pair but does not transfer to another. \emph{Test is modular; route, under these probes, is not.}

Our contributions are the \textbf{condition-vs-answer disagreement design}
itself (Figures~\ref{fig:donors},~\ref{fig:stages}), run as a shared-bank
multilingual protocol: one fixed item bank rendered fully in each of six
languages, with only the numerals and answer tokens held fixed. Applying it
yields a \textbf{separable predicate-\test{} band}, localized over two label
pairs, a three-output variant, and multi-token answers, and reproducing
across three models from two families, six languages, and five predicates;
and a \textbf{non-separable \route{}}: a learned subspace (distributed
alignment search, DAS) with a transfer control flips the trained label pair
in every model yet does not transfer to a new pair (in Gemma-3-4B it does
transfer across languages), a token-bound readout direction rather than a
reusable router; without the transfer control, the within-pair interchange
accuracy of $1.00$ would have certified a spurious router in all three
models. An exploratory contrast with the weight-stored IOI circuit
appears in Appendix~\ref{app:ioi}. Code, data, and per-cell results will be
released.\footnote{\url{https://github.com/Luxshan2000/icl-conditional-circuits}}

\section{Related Work}

\paragraph{Circuits and activation patching.}
Mechanistic-interpretability work localizes behavior to sparse subgraphs of
attention heads and MLPs by causal intervention: indirect-object
identification set the template for reading off a task-specific circuit
\citep{ioi_circuit}, activation- and path-based patching supply the
causal-mediation machinery \citep{rome,path_patching,causal_mediation}, and
circuits can be reused across tasks \citep{circuit_reuse}. These circuits are
predominantly \emph{weight-stored}: pretrained structures that a prompt merely
activates. Stage structure in in-context tasks is also emerging: a universal
mid-stack request$\to$execution split \citep{look_before_you_leap} and
modular sub-circuits for in-context propositional inference
\citep{hong_a_implies_b,syllogistic_circuits,chen_prop_logic_mech}. We
decompose an in-context \emph{conditional} (predicate evaluation vs.\
answer routing with an explicit else branch) under donors that make
condition and answer disagree.

\paragraph{In-context learning as vectors, subspaces, and functions.}
A complementary tradition compresses in-context tasks into portable
representations: function and task vectors summarize a demonstrated mapping
into a single transplantable activation \citep{function_vectors,task_vectors},
with follow-ups on their generalization, filter-like selection heads, concept
subspaces, and algorithmic primitives
\citep{function_induction_offbyone,filter_heads_lists,icl_concept_subspace,algorithmic_primitives}.
This literature asks \emph{what} mapping is applied; we ask how a conditional
rule is \emph{decomposed} into a predicate \test\ and an argument \route, and
find that under our probes the \route\ admits no separable, transferable
subspace. Transportable filter-head predicates \citep{filter_heads_lists}
independently support \test{}-separability; that setting has no else-branch
or routing analysis, where the asymmetry lies.

\paragraph{Variable binding and routing.}
How models bind entities to roles and route values to slots is central to
symbolic-style computation in transformers
\citep{binding_entities,variable_binding_symbolic}, with desiderata for
probing binding causally \citep{dcm_variable_binding_desiderata} and evidence
that models mix distinct mechanisms for a single behavior
\citep{mixing_mechanisms}. Our task is a routing problem in this sense, but
the disagreement donors expose, per layer, whether a band holds the condition
or the answer word: a \test/\route\ stage boundary rather than a single
binding operation.

\paragraph{Steering, decodability, and multilingual mechanisms.}
Directional interventions can steer behavior, even conditionally
\citep{cast_conditional_steering,refusal_single_direction}, yet a decodable or
steerable direction need not be a causally separable, invariant mechanism
\citep{steerable_not_decodable,causality_neq_invariance}, and interchange
success alone is weak evidence: optimization-found subspaces can pass it while
illusory, even on randomly initialized models
\citep{makelov_illusion,nonlinear_representation_dilemma}, and concurrent
work stress-tests head-level claims by transfer
\citep{ablation_reversible_transfer}.
Separately, multilingual analyses find computation routed through a shared,
largely English-centric internal space
\citep{llamas_work_in_english,how_llms_handle_multilingualism}, with
cross-lingual patching separating concept from language
\citep{tongue_from_thought} and near-identical circuits across two languages
\citep{circuits_across_languages}. Our transfer control distinguishes a
genuine routing subspace from answer-readout: a learned \route\ direction that
flips one label pair but fails to transfer to another is entangled with
readout, not separable. Our \test\ band recurs at mid-stack across six
languages and four scripts (including Tamil and Sinhala, to our knowledge new
to mechanistic localization), and in Gemma-3-4B the readout direction for a
fixed pair is language-invariant (\S\ref{sec:exp-router}).

\section{Methodology}
\label{sec:method}

\subsection{Task and Dataset}
\label{sec:task}
We study in-context conditional rules of the form ``if $P(x)$ then $A$ else
$B$.'' The canonical English rule is \emph{``Rule: if the number is greater
than 5, output dog, otherwise output cat.''} Each prompt concatenates the rule,
eight in-context Number/Output demonstrations, and a query number; the model
completes a single answer token. The predicate is $P(x)\!\equiv\!(x>5)$, and
the answer labels $A=\text{\emph{dog}}$, $B=\text{\emph{cat}}$ are single
tokens.

All experiments draw from \emph{one} fixed canonical bank of $250$ items,
shared by every experimental cell (every model and every language), so no
cross-cell comparison is confounded by a different random sample. The six
languages (\texttt{en}, \texttt{zh}, \texttt{hi}, \texttt{id}, \texttt{ta},
\texttt{si}) are rendered from this bank \emph{in full} (rule and
demonstration scaffold alike), while the numerals and the answer labels
\emph{dog}/\emph{cat} stay in Latin script and single-token form, so
cross-lingual tokenization variation falls on the prompt text but never on the
measured answer (Appendix~\ref{app:repro} details the translation protocol): the
manipulation targets rule and scaffold comprehension while holding the readout
interface constant. The languages span four scripts and a wide resource
range (Tamil and Sinhala are new to mechanistic localization). Each cell is
also rendered under a second single-token pair (\emph{fox}/\emph{owl});
grid curves pool both pairs.
Alternative predicate families (numeric thresholds
$t\!\in\!\{3,5,7\}$, set membership, vowel-initial) live in a separate
English-only bank used for the predicate-generality experiment.

\subsection{Activation Patching and Metrics}
\label{sec:patching}
We localize computation with activation patching
\citep{causal_mediation,rome,path_patching}: run the model on a \basecond{}
prompt and on a donor prompt, copy the donor's residual-stream vector at a
single layer, at the last-token position, into the \basecond{} run, and
measure whether the predicted answer changes; repeated per layer, this
localizes which band carries which piece of the computation.

\begin{figure}[t]
  \centering
  \includegraphics[width=\columnwidth]{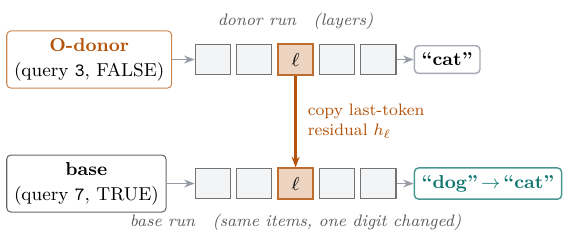}
  \caption{The patching operation. The donor's last-token residual $h_\ell$ is
  copied into the \basecond{} run at layer $\ell$. We then record, at every
  layer, whether the patched answer flips (under the label-mass gate,
  $\ge 0.30$) and the graded fraction-of-swing recovery $R$.}
  \label{fig:patching}
\end{figure}

We report two metrics \citep{patching_best_practices,patching_howto}. The
first is a two-way \emph{flip probability} over
$\{A,B\}$, gated by label mass: a patch counts as a flip only if the combined
probability mass on $\{A,B\}$ is at least $0.30$, so destructive patches that
collapse the distribution do not register as flips. The second is a graded
\emph{fraction-of-swing} recovery
\[
  R \;=\; \frac{\mathrm{LD}_{\text{patched}} - \mathrm{LD}_{\text{base}}}
               {\mathrm{LD}_{\text{donor}} - \mathrm{LD}_{\text{base}}},
\]
where $\mathrm{LD} = \mathrm{logit}(A) - \mathrm{logit}(B)$: how far a patch
moves the base logit difference toward the donor's ($R\!=\!0$ no movement,
$R\!=\!1$ full recovery). When this fraction-of-swing is computed for a
patched head group or subspace rather than a full-layer residual patch, we
write it as $\Phi$. Both flip directions are always run (predicate
\textsc{true}$\to$flip and \textsc{false}$\to$flip);
Figure~\ref{fig:patching} illustrates the operation.

\subsection{The Four Donors}
\label{sec:donors}
\begin{figure}[t]
  \centering
  \includegraphics[width=\columnwidth]{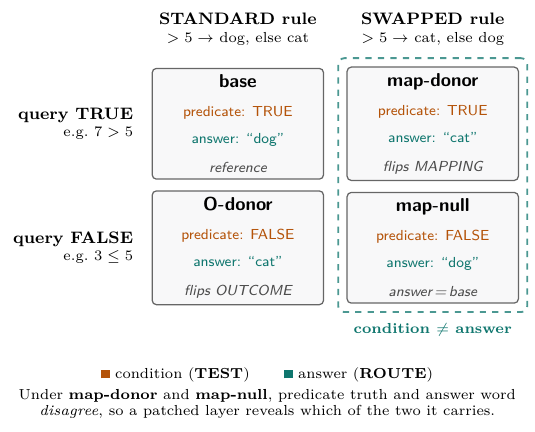}
  \caption{The four donors, crossing rule mapping (standard vs.\ swapped) with
  query outcome (\textsc{true} vs.\ \textsc{false}). In \mapdonor{} and
  \mapnull{} the predicate truth and the answer word \emph{disagree}, exposing
  per layer whether a patched site follows the condition (\test{}) or the answer
  (\route{}).}
  \label{fig:donors}
\end{figure}

For each \basecond{} item (predicate \textsc{true}, answer
$A=\text{\emph{dog}}$) we construct four prompts by crossing the rule
\emph{mapping} (standard vs.\ swapped) with the query \emph{outcome}
(\textsc{true} vs.\ \textsc{false}), as laid out in Figure~\ref{fig:donors}:
\begin{itemize}
  \setlength{\itemsep}{1pt}
  \item \basecond{}: standard rule, query \textsc{true} $\Rightarrow$ predicate
    \textsc{true}, answer \emph{dog} (reference run).
  \item \odonor{}: standard rule, query \textsc{false} $\Rightarrow$ predicate
    \textsc{false}, answer \emph{cat}. Flips the predicate \emph{outcome}; only
    one query digit changes.
  \item \mapdonor{}: swapped rule (``\dots output cat, otherwise dog''), query
    \textsc{true} $\Rightarrow$ predicate stays \textsc{true} but the answer
    becomes \emph{cat}. Flips the \emph{mapping}.
  \item \mapnull{}: swapped rule, query \textsc{false} $\Rightarrow$ predicate
    \textsc{false}, answer \emph{dog}. A large perturbation whose answer
    nonetheless equals \basecond{}.
\end{itemize}
The methodological core is that \mapdonor{} and \mapnull{} make the
\emph{condition} and the \emph{answer} \emph{disagree} (truth value and
answer word point to opposite labels), letting us read, per layer, a patch's
\emph{allegiance}: whether the patched output follows the truth value (the
\test{} stage) or the answer word (the \route{} stage). The construction is
binary here for clarity, but generalizes: a cyclic rotation of the label
mapping extends it to a three-output rule (\S\ref{sec:exp-multilabel}).

\subsection{Controls}
\label{sec:controls}
A \emph{behavioral gate} admits a cell only if base accuracy is at least
$0.85$, so that the model actually solves the rule before we interpret its
internals. \emph{Donor-correctness gating} discards any item whose donor
prompts are not themselves answered correctly. A \emph{perturbation-matched
$2\times2$ isolation} test contrasts \odonor{} (outcome-flipping, one digit
changed) against \mapdonor{} (mapping-flipping, predicate held), certifying
that a \test{}-band patch reroutes the answer by moving the \emph{condition}
rather than by any incidental surface perturbation.

To rule out a subspace illusion, an \emph{illusion-guard} compares recovery
along the learned truth direction against a random direction of equal norm
(values in \S\ref{sec:exp-test}). For the argument \route{} we add a \emph{cross-label-pair transfer
control}, applying a subspace learned on \emph{dog}/\emph{cat} to a held-out
label pair; failure to transfer marks the router as entangled with readout
rather than separable. We further verify a valid \emph{coherent null}
(\textsc{null\_cross}: the residual of an unrelated base item) and confirm
that an ungated vector null (\textsc{null\_vec}: a norm-matched random
vector) collapses once the label-mass gate is applied, so apparent flips are
not artifacts of distributional collapse. Significance is
assessed with McNemar's test on paired flip outcomes, each cell reporting a
single run over its full gated item set.

\paragraph{Verdict criteria.}
All thresholds were fixed before the grid was run: behavioral gate at accuracy
$\ge 0.85$; analysis restricted to items whose four donor prompts are all
answered correctly; \test{} layer taken as the argmax of the \odonor{} flip
curve; and strict single-layer isolation requiring, at that layer, \mapnull{}
flip $\ge 0.6$ with \mapdonor{} flip $\le 0.2$. Per-cell values appear in
Table~\ref{tab:e1}, full curves in Appendix~\ref{app:curves}.

\section{Experiments and Results}
\label{sec:experiments}

Appendix Figure~\ref{fig:design} summarizes the design:
\emph{predicate-\test{} localization} (\S\ref{sec:exp-test}) in every cell
of a $6$-language $\times$ $3$-model grid over the shared bank; the
\emph{router analysis} (\S\ref{sec:exp-router}) on all three models;
generality across families and languages (\S\ref{sec:exp-generality}); and
\emph{predicate generality} (\S\ref{sec:exp-e3}) on the primary model. A recruit-vs-build contrast with the weight-stored IOI
circuit is deferred to Appendix~\ref{app:ioi}. A language enters an analysis only
if the model's base accuracy on the shared bank is at least $0.85$ (the
behavioral gate); every cell reported below clears it ($[0.926, 1.000]$).

\subsection{Localizing the Predicate Test}
\label{sec:exp-test}

For each item we run per-layer residual patching of the four donors
(\basecond, \odonor, \mapdonor, \mapnull) at the last-token position. Because
\mapdonor{} and \mapnull{} make condition and answer \emph{disagree}, the
patched prediction reveals, per layer, whether that layer carries the
condition (\test{}) or the routed answer (\route{}). The primary metric is the $2$-way flip probability under
the label-mass gate, complemented by the graded recovery $R$
(\S\ref{sec:patching}). Every curve pools two single-token answer pairs
(\emph{dog}/\emph{cat}, \emph{fox}/\emph{owl}) and both flip directions.

\begin{figure*}[t]
  \centering
  \includegraphics[width=\textwidth]{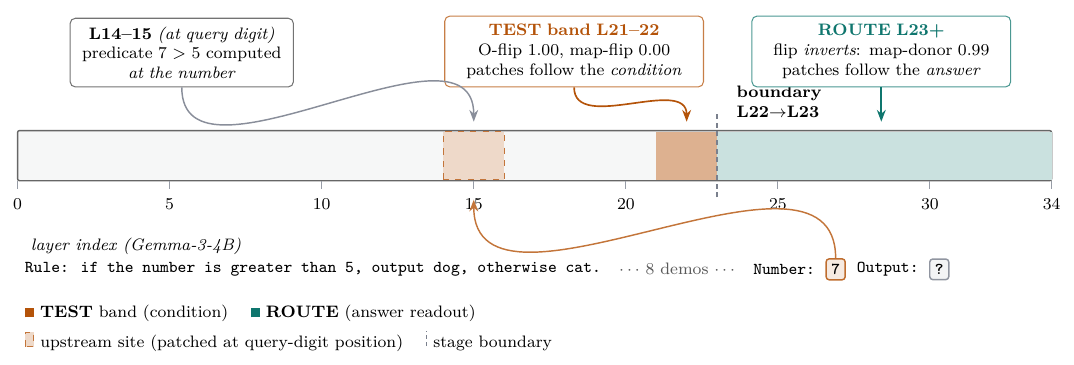}
  \caption{The \test{}$\rightarrow$\route{} stage boundary in Gemma-3-4B
  (English). All patches are applied at the last token of the prompt (the
  \texttt{Output:} position), which is also the readout site. The predicate is
  first computed at the query-digit position
  (L14--15), read out as a truth value at the mid-stack \test{} band (L21--22,
  amber), where the \odonor{} reroutes the answer ($1.00$) but the \mapdonor{}
  does not ($0.00$); patches there follow the \emph{condition}
  (\mapdonor{}~$0.00$, \mapnull{}~$\approx1.0$). At L23 the pattern
  inverts (teal): patches follow the \emph{answer word}, marking the
  answer-readout \route{} stage. Gemma-3-4B is the \emph{fused} case:
  readout begins immediately past the band; see
  Figure~\ref{fig:bandpeaks} for the staggered geometries of Gemma-3-12B
  (readout onset L29--L35, language-dependent) and Qwen3-8B (L30).}
  \label{fig:stages}
\end{figure*}

\paragraph{A separable, isolated \test{} module.}
Table~\ref{tab:e1} reports the complete 18-cell grid; every cell but one
returns \textsc{pass\_test\_isolated} (the exception, Qwen3-8B Tamil, is a
layer-selection artifact; \S\ref{sec:exp-generality}): at a single mid-stack
residual band, the \odonor{}, which flips only the predicate's truth value, reroutes
the answer with probability $\approx 1.0$, while the \mapdonor{}, which flips
the rule's mapping but leaves the predicate \textsc{true}, leaves the answer
essentially unchanged ($\approx 0.0$). The
McNemar tests reject the null throughout ($p$ between $1.4\times10^{-47}$ and
$5.7\times10^{-57}$).

\begin{table}[t]
\centering
\small
\setlength{\tabcolsep}{2.5pt}
\begin{tabular}{@{}lllccccc@{}}
\toprule
Lang & Script & Model & Acc & \test{} & peak & O/map & $p$ \\
\midrule
en & Latin & 4B   & 0.961 & 22/34 & 22 & 1.00/0.00 & 3.6e{-}52 \\
en & Latin & 12B  & 0.996 & 27/48 & 27 & 0.99/0.00 & 1.9e{-}55 \\
en & Latin & Qwen & 1.000 & 27/36 & 23 & 0.99/0.00 & 1.6e{-}56 \\
\midrule
zh & Han   & 4B   & 0.949 & 21/34 & 21 & 0.99/0.01 & 2.5e{-}49 \\
zh & Han   & 12B  & 0.981 & 27/48 & 27 & 0.99/0.00 & 2.4e{-}54 \\
zh & Han   & Qwen & 1.000 & 23/36 & 23 & 1.00/0.00 & 5.7e{-}57 \\
\midrule
hi & Deva  & 4B   & 0.953 & 21/34 & 22 & 1.00/0.00 & 6.7e{-}49 \\
hi & Deva  & 12B  & 1.000 & 27/48 & 27 & 0.97/0.00 & 1.9e{-}55 \\
hi & Deva  & Qwen & 0.965 & 25/36 & 23 & 1.00/0.00 & 2.2e{-}52 \\
\midrule
id & Latin & 4B   & 0.938 & 21/34 & 21 & 0.99/0.01 & 1.8e{-}48 \\
id & Latin & 12B  & 1.000 & 28/48 & 27 & 0.96/0.00 & 8.7e{-}55 \\
id & Latin & Qwen & 0.992 & 25/36 & 23 & 1.00/0.00 & 2.6e{-}56 \\
\midrule
ta & Tamil & 4B   & 0.938 & 22/34 & 22 & 0.99/0.02 & 8.3e{-}48 \\
ta & Tamil & 12B  & 0.984 & 27/48 & 27 & 0.96/0.01 & 5.9e{-}52 \\
ta & Tamil & Qwen$^\dagger$ & 1.000 & 25/36 & 23 & 1.00/0.00 & 5.7e{-}57 \\
\midrule
si & Sinh. & 4B   & 0.926 & 22/34 & 22 & 1.00/0.00 & 1.4e{-}47 \\
si & Sinh. & 12B  & 0.984 & 28/48 & 27 & 0.96/0.01 & 4.8e{-}53 \\
si & Sinh. & Qwen & 0.996 & 23/36 & 23 & 0.99/0.00 & 5.3e{-}55 \\
\bottomrule
\end{tabular}
\caption{Predicate-\test{} localization across the 18-cell grid. Columns:
base accuracy (all clear the $0.85$ gate); \test{}, the
\odonor{}-selected layer over model depth (e.g.\ $22/34$: layer 22 of 34);
\emph{peak}, the condition peak (\mapnull{} argmax); O/map, the
\odonor{}/\mapdonor{} flip probabilities; McNemar $p$. Gated $n$ per cell
ranges $214$--$256$. Models: Gemma-3-4B,
Gemma-3-12B, Qwen3-8B. Every cell but one ($^\dagger$) returns
\textsc{pass\_test\_isolated}. $^\dagger$At the \odonor{}-selected layer the
\mapnull{} flip is $0.57$, just under the $0.6$ isolation cutoff: an
artifact of the layer rule, not a Tamil effect; at the condition peak it
passes, as every Qwen cell does (\S\ref{sec:exp-generality}).}
\label{tab:e1}
\end{table}

\paragraph{Stage boundary and controls.}
Figure~\ref{fig:stages} shows the stage structure on the primary cell
(English, Gemma-3-4B). Three controls confirm a directed truth-value edit, not
an artifact. (i) \emph{Subspace illusion-guard}: patching along the learned
truth direction recovers the swing at $0.67$, versus $0.03$ for a random
direction of equal norm. (ii) \emph{Upstream localization}: patching the
query-digit token position flips the answer most strongly at L14--15
($0.83$/$0.75$), decaying to zero by L20: the predicate is first computed
\emph{at} the number and only later moved to the last token. (iii) \emph{Stage
boundary at L22$\to$L23}: within the band (L20--22) patches follow the
\emph{condition} (\mapdonor{} $0.00$, \mapnull{} $0.97$--$1.00$); from L23
the pattern \emph{inverts} (\mapdonor{} $0.99$, \mapnull{} $0.00$), marking
the transition into answer readout. The coherent null
(\textsc{null\_cross}) stays at $\approx 0.04$, and \textsc{null\_vec} at
$\approx 0.02$ once the label-mass gate is applied.

\subsection{Is the Router Separable?}
\label{sec:exp-router}

\begin{figure}[t]
  \centering
  \includegraphics[width=\columnwidth]{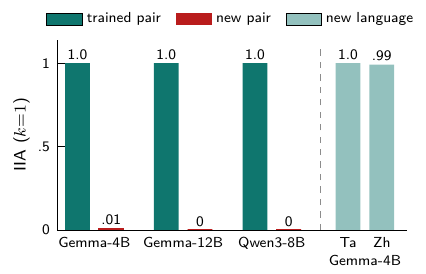}
  \caption{The router transfer pattern ($k{=}1$). In every model the learned
  subspace flips the trained pair perfectly (teal) but transfers to a new pair
  at ${\approx}0$ (red); for Gemma-3-4B the same subspace transfers
  near-perfectly to the \emph{same} pair in Tamil and Chinese (light teal):
  token-bound, language-invariant.}
  \label{fig:e2transfer}
\end{figure}

Does the \route{} step (mapping the truth value to the answer word) have its
own separable, transferable subspace? The readout onset (the first layer
where the \mapdonor{} flip reaches $0.9$) differs by model (Appendix
Figure~\ref{fig:bandpeaks}), so each probe is placed at that model's
\emph{own} onset (Table~\ref{tab:e2}). Method 1 greedily accumulates attention heads over each
model's readout band (4B L23--30; 12B L32--43; Qwen L30--35). Method 2, ``DAS-lite'', is a low-rank variant of distributed
alignment search \citep{das,boundless_das}: a rank-$k$ orthogonal basis at
one residual site, trained on the interchange objective and scored by
interchange intervention accuracy (IIA) on held-out items
($k\in\{1,2,4,8\}$, $48/32$ train/test). Both methods run on all three
models (English); Appendix Figure~\ref{fig:e2} illustrates them on the
primary cell, Figure~\ref{fig:e2transfer} the transfer pattern,
Table~\ref{tab:e2} all three models. These probes span greedy head
accumulation \citep{acdc} over each full readout band and rank-$1$--$8$
subspaces at the onset site, tested cross-pair in every model,
cross-lingually in Gemma-3-4B, and under \emph{joint} two-pair training; a
subspace anywhere in this space that transferred across label pairs would
overturn the claim, and none does. The decisive joint case (Gemma-3-4B): a
subspace trained on \texttt{dog}/\texttt{cat} and \texttt{block}/\texttt{allow}
together fits both at $\mathrm{IIA}=1.00$ down to rank $1$ yet transfers at
$\le0.01$ to held-out \texttt{fox}/\texttt{owl}; its only partial transfer
($\le0.46$, rank-unstable) is to \texttt{flag}/\texttt{approve}, lexically
parallel to a trained pair: token geometry, not routing
(Appendix~\ref{app:repro}).

\paragraph{Greedy head-group patching.}
Where this probe recovers the routing effect at all, the \mapdonor{}
(mapping) and \odonor{} (outcome) donors recruit the \emph{same} heads in
nearly the same greedy order: identical for the first
five of six heads in Gemma-3-4B (L23H3, L23H1, L30H6, L29H4, L29H5; $0.8\times$
of the full-band swing with $3$ heads, full-band $\Phi=0.96$), and identical
for \emph{all six} in Qwen3-8B (L33H9, L35H26, L31H6, L30H5, \dots; full-band
$\Phi=0.58$). Gemma-3-12B differs informatively: patching \emph{all} band
heads at once is nearly null (full-band $\Phi=0.09$), yet the greedy-selected
ten-head group recovers $\Phi=0.58$ with flips $0.91$, led by the same head
(L41H11) for both donors, nine of ten heads shared. In no model do distinct
routing heads emerge: mapping and outcome share the same machinery, their
accumulation curves nearly coincident (Appendix Figure~\ref{fig:e2greedy}).

\paragraph{A learned subspace with a transfer control.}
At each model's readout layer, a learned low-rank subspace (DAS-lite) flips
\texttt{dog}\,$\leftrightarrow$\,\texttt{cat} with interchange accuracy
$\mathrm{IIA}=1.00$ on held-out items at every rank $k\in\{1,2,4,8\}$
(Table~\ref{tab:e2}); yet under the decisive \emph{cross-label-pair transfer
control}, the same subspace applied to a fresh label pair
(\texttt{fox}/\texttt{owl}) flips at $\le 0.013$ in every model ($0.00$ for
Gemma-3-12B and Qwen3-8B, $n\ge 462$). The subspace's alignment with the
answer-readout direction (the normalized mean \mapdonor{}$-$\basecond{}
residual difference; largest $|\cos|$ over learned columns), however,
\emph{differs} by family: in the Gemma
models it is moderate (cosine $0.15$--$0.39$), so the flip direction is largely
the readout direction itself; in Qwen3-8B it is near zero (cosine
$0.01$--$0.09$), a \emph{pair-specific} direction that neither transfers nor
coincides with readout. Optimization-found subspaces can pass interchange
tests while illusory \citep{makelov_illusion,nonlinear_representation_dilemma};
the transfer control is the direct antidote.

\paragraph{Cross-lingual transfer: pair-specific, not language-specific.}
A second transfer probe pins down what the direction \emph{is}: we apply the
English-learned \texttt{dog}/\texttt{cat} subspace (Gemma-3-4B, L23),
\emph{unchanged}, to the same pair in another language (only the rule text
differs; the labels are shared Latin tokens). Whereas transfer to a new
label pair fails ($\le 0.011$), transfer to Tamil and Chinese \emph{succeeds}
almost perfectly at every rank (Tamil $0.97$--$1.00$, Chinese $0.99$; the
greedy probe re-run in this setting recruits the same heads). The router
direction is thus \emph{token-specific but language-invariant}: the
answer-readout direction for those particular tokens, shared across languages
because the tokens are; that token-binding, not any language
binding, makes it non-transferable.

\begin{table}[t]
\centering
\small
\setlength{\tabcolsep}{3.1pt}
\begin{tabular}{@{}lccccc@{}}
\toprule
Model & DAS & $\mathrm{IIA}_{\text{test}}$ & transfer & cos(readout) & greedy \\
 & layer & $k{=}1{:}8$ & fox/owl & $k{=}1{:}8$ & $\Phi_{\text{full}}$ \\
\midrule
Gemma-4B  & L23 & 1.00 & $\le$0.013 & 0.27--0.39 & 0.96 \\
Gemma-12B & L32 & 1.00 & 0.00 & 0.15--0.37 & 0.09 \\
Qwen-8B   & L30 & 1.00 & 0.00 & 0.01--0.09 & 0.58 \\
\bottomrule
\end{tabular}
\caption{Router analysis across three models spanning two families (English). DAS flips
the answer perfectly at every rank ($\mathrm{IIA}_{\text{test}}=1.00$) but
\emph{transfers to a new label pair at $\approx 0$} in every model: no
separable, transferable router. Cosine with the readout direction is moderate
in Gemma (the flip is largely readout) but near-zero in Qwen (a pair-specific
direction). The greedy head-patch recovers the swing in 4B/Qwen but not 12B
(full-band $\Phi$), so the 12B row rests on the subspace method alone.
$\mathrm{IIA}$ is on $32$ held-out items; transfer is over $n\ge462$ gated
items. A larger-sample rerun of the primary cell ($150$ train, $100$ test)
gives $\mathrm{IIA}_{\text{test}}=0.98$--$0.99$ across ranks with transfer
$\le0.011$, so the interchange result is not a small-sample effect.}
\label{tab:e2}
\end{table}

\subsection{Generality Across Model Families and Languages}
\label{sec:exp-generality}

Table~\ref{tab:e1} is itself the cross-cell evidence against three
worries. \emph{Model idiosyncrasy}: the decomposition replicates
across Gemma-3-4B, Gemma-3-12B \citep{gemma3}, and Qwen3-8B \citep{qwen3}
(the cross-\emph{family} comparison is the real test), with
the \test{} band at comparable relative depth in each (Gemma-3-4B L21--22,
$\sim\!0.63$; Gemma-3-12B L27--28, $\sim\!0.57$; Qwen3-8B L23--27).
\emph{Lexical specificity}: beyond the two pooled pairs of
\S\ref{sec:exp-test}, English reruns with policy-\emph{action} pairs
(\emph{block}/\emph{allow}, \emph{flag}/\emph{approve}) and with
\emph{multi-token} answers (sub-word splits and two-word phrases;
sequence-margin metrics) reproduce the band, isolation, and boundary
(Appendix~\ref{app:realistic}); the band tracks the predicate's truth
value, not a particular answer word or token count.
\emph{Language specificity}: across six languages spanning four scripts, the
\test{} band sits at the same mid-stack depth with the same isolation signature
(O-flip $\approx 1.0$, map-flip $\approx 0.0$). Both sides are
language-agnostic in different senses: the predicate test localizes at a
shared depth, and the readout \emph{direction} is language-invariant (the
cross-lingual probe above transfers at $\approx 0.98$); only the readout
\emph{onset} can shift by language (12B, Figure~\ref{fig:bandpeaks}), consistent
with reports of shared mid-stack multilingual computation
\citep{llamas_work_in_english, how_llms_handle_multilingualism,
transfer_neurons}.

\paragraph{A cross-family difference in band structure.}
The per-layer curves expose three per-cell quantities
(Appendix Figure~\ref{fig:bandpeaks}): the \emph{condition peak} (\mapnull{} argmax),
the selected \test{} layer (\odonor{} site), and the \emph{readout onset}
(\S\ref{sec:exp-router}). Gemma-3-4B is \emph{fused}: all
three sit within two layers in every language (onset L23 in all six).
Qwen3-8B has \emph{two language-invariant anchors}: condition peak L23 and
readout onset L30 in \emph{all six} languages, the selected \test{} layer
varying between them. This exposes a limit of our
pre-registered layer rule: the \odonor{} flips condition \emph{and} answer, so
its curve saturates across the corridor between condition peak and readout
onset (Qwen $\ge\!0.95$ from L23 on) and the
argmax is settled by single-item differences (Table~\ref{tab:e1},
\test{} vs.\ peak). Where the argmax lands past L23, \mapnull{} has
attenuated and Tamil dips below the $0.6$ cutoff; at the condition peak
\emph{all six} meet the criterion. The pre-registered verdicts in
Table~\ref{tab:e1} stand. Gemma-3-12B anchors the condition at
L27 everywhere but its \emph{readout onset is language-dependent} (zh L29;
en, hi, ta L32; id, si L35). The same attenuation marks every
corridor of two or more layers (corridor-mean \mapnull{} flips of
$0.00$--$0.55$)
while the matched-rule \odonor{} still flips, a graded
condition$\to$readout transition rather than a single boundary. Throughout, the
\test{} anchor is language-invariant; only \emph{where} the readout
crystallizes varies, not its content (\S\ref{sec:exp-router}).

\subsection{Predicate Generality}
\label{sec:exp-e3}

Is the mechanism specific to greater-than-$5$, or a general predicate-test? We
hold the language fixed (English) and vary the predicate family, keeping the
same format and \texttt{dog}/\texttt{cat} labels so only the predicate
changes: numeric thresholds $t{\in}\{3,5,7\}$, set-membership
$x\!\in\!\{2,3,5,7\}$ (a non-monotonic operator), and vowel-initial (a lexical
predicate on letters, not numbers). For each family we localize the \test{}
with the per-layer outcome-donor patch against a coherent null.
Appendix Table~\ref{tab:e3} shows that \emph{every}
family exhibits a clean mid-stack \test{} band with peak outcome-flip $1.00$
and near-zero null, including the non-monotonic set-membership and the
non-numeric vowel predicate; a Qwen3-8B spot-check on the two hardest families
(set-membership, vowel) reproduces this (peak-flip $1.00$ at L28 and L30). The
mechanism is therefore a \emph{predicate-test}, not a greater-than-$5$ circuit
(per-layer curves in Appendix Figure~\ref{fig:e3gen}). Selected \test{}
layers vary within the band (L20--L29), but every family reaches flip
$\ge 0.97$ from L21 on, including the two whose accuracy sits near the
behavioral gate.

\subsection{Beyond Binary Rules}
\label{sec:exp-multilabel}

We ran a \emph{three-output}
variant on Gemma-3-4B (English): three numeric buckets map to three
single-token labels (\emph{dog}/\emph{cat}/\emph{cow}), and the four donors
generalize by a cyclic rotation of the label mapping in place of the binary
swap (construction in Appendix~\ref{app:multilabel}). The result matches the
binary case: the model
solves the rule (behavioral accuracy $0.91$) and the \emph{same} mid-stack
\test{} band reappears (L18--22, stage boundary L22$\to$L23), with the same
isolation signature: \odonor{} and \mapnull{} flip ($1.00$/$0.98$)
while \mapdonor{} does not ($0.00$)
(Appendix Figure~\ref{fig:ml3}): the predicate-\test{} module is not tied
to a two-label task.

\section{Discussion}

Our central finding is an asymmetry: a single mid-stack band carries the
rule's truth value and can be overwritten in isolation, across three models
and six languages, while under our probes the \route{} has no such handle;
its learned direction flips only the trained pair.

The \test{} stage behaves like a computation over an abstract truth
value: the rule text sets up a predicate, evaluated at the query digit and
moved to the last token, its result carried on a direction stable enough to swap
between prompts. Because outputs follow the truth value below the boundary and the answer
above it, the disagreement design licenses calling the \test{} a module
rather than a shortcut. Two controls argue the band carries the evaluated
truth value rather than a relocated query digit (the digit's causal weight
decays to zero by L20; a learned truth direction alone recovers $0.67$ of
the swing), though a cross-threshold patch, not run here, would settle it. The condition anchor, moreover, is
language-invariant in every family (Appendix Figure~\ref{fig:bandpeaks}),
as expected of a language-agnostic truth value; only the readout's
\emph{timing} varies by family and, in 12B, by language.

The \route{}, though, shows no separable, transferable handle. In all three
models a DAS-lite subspace attains IIA $1.00$ yet fails to transfer to a new
label pair ($\approx 0$; Table~\ref{tab:e2}). What that direction \emph{is}
varies by family: in Gemma it aligns moderately with answer readout, the
apparent router largely the readout seen through one label pair; in Qwen it
is orthogonal to readout, a pair-specific direction; the invariant is
non-transferability. The cross-lingual probe pins the binding: failing on a new pair yet
transferring across languages, the direction tracks the answer
\emph{tokens}, not the rule's language or a routing role. This
matches reports of directions that steer effectively without being a
faithful, transferable code
\citep{steerable_not_decodable,causality_neq_invariance}.

Practically, the asymmetry predicts that predicate-level steering should be
more portable across prompts and label sets than answer-level steering.
In the settings where such conditionals ship
(Appendix~\ref{app:realistic}), single-direction steering
\citep{cast_conditional_steering,refusal_single_direction} may therefore
generalize unevenly, surviving a change of \emph{language} but not
\emph{vocabulary}; the transfer test cheaply tells the two apart.
A separable \test{} also offers a
monitoring hook: the judged condition is readable mid-stack \emph{before}
the action is emitted, a causal target for truth-probing and
judgment/action-mismatch detection
\citep{latent_knowledge,geometry_of_truth}.

Holding the answer tokens fixed in Latin script is a deliberate control:
native-script labels would make the readout select \emph{and} render the
answer, confounding routing with translation.

\section{Conclusion}

In-context conditional execution factors asymmetrically. The predicate
\test{} is a separable mid-stack module carrying the rule's truth value:
localized by disagreement donors, isolated under matched controls, stable
across three models and six languages (on Gemma-3-4B also five predicates,
a three-output rule, and multi-token answers). No probe, single-pair or
joint, found a transferable \route{} subspace: the direction is
answer-token-bound, failing new label pairs while transferring across
languages in Gemma-3-4B.

\section*{Limitations}
Our results come with clear boundaries. All experiments use three open-weight
models (Gemma-3-4B/12B, Qwen3-8B); we cannot patch proprietary systems, so we
cannot say whether the \test{}/\route{} asymmetry persists at frontier scale.
The task is a deliberately controlled synthetic conditional with
\emph{dog}/\emph{cat} labels held in Latin script across languages: this is
what makes the flip metrics interpretable and the readout direction shareable
across languages. A multi-token rerun on the primary cell reproduces the
localization under sequence-margin metrics, but native-script conditionals
and free-form completions remain open. Activation patching
localizes causally load-bearing sites; our claims concern localization and
separability, not a complete circuit-level mechanism. On the router side, the
full-band head patch is nearly null in Gemma-3-12B ($\Phi=0.09$; a greedy
ten-head subset recovers $0.58$), so the 12B transfer negative rests on the
subspace method; the full two-method analysis is English-only, and the
cross-lingual transfer probe was run on Gemma-3-4B only (per-language
subspace training was not repeated); ``entangled with readout'' is precise
for Gemma while Qwen's direction is pair-specific without aligning to
readout. Joint two-pair training (\S\ref{sec:exp-router}) closes the
single-pair loophole on the primary model, but wider multi-pair banks, other
sites, and nonlinear alignments remain unprobed. Lexical generality rests on two label pairs in the grid (plus two
policy-action pairs and two multi-token pairs on the primary cell), the three-output variant was run on
Gemma-3-4B in English only, and each cell reports a single seed. The IOI contrast
(Appendix~\ref{app:ioi}) is suggestive rather than decisive, since the IOI
movers are themselves weak and distributed in Gemma-3-4B. Finally, the six
languages sample script and resource level, not typological space.

\section*{Ethics Statement}
This work is a mechanistic-interpretability study of open-weight language models (Gemma-3-4B, Gemma-3-12B, and Qwen3-8B) on a synthetic task. It involves no human subjects, no user data, and no private or personally identifiable information: every stimulus is generated deterministically from a fixed numeric item bank and a rule template. The multilingual stimuli are machine-rendered from the same template into six languages, with the rule and demonstration scaffold translated and the single-token answer labels held in Latin; the translated rule text was checked by native speakers for each language to ensure the intended predicate is faithfully expressed. All models are publicly released, and our interventions (activation patching on the residual stream) modify only internal activations at inference time, leaving model weights unchanged.

We note a dual-use consideration. Localizing a causally separable predicate-\test\ module identifies a residual-stream direction at which flipping the truth value reroutes the answer, and such localization could in principle inform activation-steering attempts to manipulate a model's conditional behavior. We regard the risk as limited: our methods require white-box access to internal activations, the studied task is a controlled synthetic conditional rather than a safety-relevant behavior, and the same interpretability tools are what make such manipulations auditable and detectable. We report these findings to advance scientific understanding of how in-context rules are assembled into runtime circuits, and we encourage that steering insights derived from interpretability be applied toward transparency and control rather than covert behavioral manipulation.


\bibliography{references,references_extra}

\appendix
\section{Reproducibility Details}
\label{app:repro}

\paragraph{Prompt template.}
Each prompt is a rule, eight in-context Number/Output demonstrations, and a query
number whose answer token the model completes. The canonical English rule is
``Rule: if the number is greater than $5$, output dog, otherwise output cat.''
Answer labels \emph{dog}/\emph{cat} are single tokens and stay in Latin script across
all six languages (en, zh, hi, id, ta, si); the rule and the entire
demonstration scaffold are translated, so tokenization variation falls on the
prompt text and never on the measured answer.
Figure~\ref{fig:prompts} shows one English prompt per predicate family (each
abbreviated to two of eight demonstrations). The multilingual banks were built
by translating this template (rule text and scaffold keywords) into each
of the other five languages and having the result checked and corrected by a
native speaker, keeping the numerals and the answer tokens unchanged.

\begin{figure}[!ht]
\centering
\small
\begin{tabular}{@{}l@{}}
\toprule
\textbf{Numeric threshold} (also $t{=}3$, $t{=}7$) \\
\midrule
\texttt{Rule: if the number is greater than 5,} \\
\texttt{\ \ output dog, otherwise output cat.} \\
\texttt{Number: 0} \quad \texttt{Output: cat} \\
\texttt{Number: 6} \quad \texttt{Output: dog} \\
\texttt{\ldots{} (8 demonstrations) \ldots} \\
\texttt{Number: 7} \quad \texttt{Output:} \\
\midrule
\textbf{Set membership} \\
\midrule
\texttt{Rule: if the number is in the set} \\
\texttt{\ \ \{2, 3, 5, 7\}, output dog, otherwise cat.} \\
\texttt{Number: 4} \quad \texttt{Output: cat} \\
\texttt{Number: 2} \quad \texttt{Output: dog} \\
\midrule
\textbf{Vowel (categorical, letters)} \\
\midrule
\texttt{Rule: if the letter is a vowel,} \\
\texttt{\ \ output dog, otherwise output cat.} \\
\texttt{Letter: b} \quad \texttt{Output: cat} \\
\texttt{Letter: a} \quad \texttt{Output: dog} \\
\bottomrule
\end{tabular}
\caption{English prompts for each predicate family (abbreviated to two of
eight demonstrations). Only the predicate changes; the task format and the
single-token \texttt{dog}/\texttt{cat} labels are held fixed. Non-English
banks translate the rule and scaffold keywords, with native-speaker
correction, leaving numerals and answer tokens unchanged.}
\label{fig:prompts}
\end{figure}

\paragraph{Fixed item bank.}
Every experimental cell (each model $\times$ each language) draws from one shared bank
of $250$ items with the numeric greater-than predicate at threshold $5$; the shared bank ensures no cell is confounded by different random samples.
Predicate families for the generality analysis live in a separate English-only bank; both banks ship with the code release.

\paragraph{Four-donor construction.}
The four donors cross rule mapping (standard/swapped) with query outcome
(\textsc{true}/\textsc{false}) exactly as defined in \S\ref{sec:donors};
we do not repeat them here.

\paragraph{Models and precision.}
Gemma-3-4B ($34$ layers), Gemma-3-12B ($48$ layers), and Qwen3-8B ($36$ layers), all run
in \texttt{bfloat16}. Layers are $0$-indexed over decoder blocks; a patch at layer $\ell$
copies the donor's residual-stream vector at the block-$\ell$ output, at the last token
position, into the base run.

\paragraph{Metrics.}
After patching one layer we score the completion. The flip probability is a $2$-way mass
over $\{A,B\}$ with a label-mass gate: a patch counts only when the combined label mass is
$\geq 0.30$, so destructive or incoherent patches do not register as flips. The graded
fraction-of-swing recovery is
\[
  R = \frac{\mathrm{LD}_{\text{patched}} - \mathrm{LD}_{\text{base}}}
           {\mathrm{LD}_{\text{donor}} - \mathrm{LD}_{\text{base}}},
\]
with $\mathrm{LD} = \mathrm{logit}(A) - \mathrm{logit}(B)$.
Both flip directions (\textsc{true}$\rightarrow$flip and \textsc{false}$\rightarrow$flip) are run.

\paragraph{DAS-lite training.}
The subspace is a semi-orthogonal basis $Q\in\mathbb{R}^{d\times k}$
(orthogonal parametrization), trained with Adam (learning rate $10^{-2}$,
batch $16$, $150$ epochs) to minimize the cross-entropy of the patched
logits against the counterfactual mapping target; IIA is
argmax-among-labels agreement with that target on held-out items, and the
null is an untrained random subspace of matched rank. Every experiment
reports a single seed per cell.

\paragraph{Joint two-pair DAS.}
Single-pair training cannot in principle discover a router shared across
pairs, so on the primary cell (English, Gemma-3-4B) we retrained the
subspace jointly on \texttt{dog}/\texttt{cat} and
\texttt{block}/\texttt{allow} (interleaved batches, $24$ train/$16$ test per
pair, same site, objective, and gates). At every rank $k\in\{1,2,4,8\}$ the
joint subspace reaches $\mathrm{IIA}_{\text{test}}=1.00$ on \emph{both}
trained pairs (rank $1$ suffices for both jointly), yet transfers at
$\le0.009$ to the unrelated held-out pair \texttt{fox}/\texttt{owl}
($n{=}114$; random-subspace null $0.00$). Transfer to
\texttt{flag}/\texttt{approve} ($n{=}112$) is partial and rank-unstable
($0.00$, $0.46$, $0.07$, $0.32$ for $k{=}1,2,4,8$): what transfer exists
follows the lexical similarity of the answer tokens
(\texttt{block}${\sim}$\texttt{flag}, \texttt{allow}${\sim}$\texttt{approve})
and vanishes for unrelated vocabulary, the signature of readout-geometry
overlap rather than an abstract routing variable.

\paragraph{Grid layout.}
In the released code, results are organized per language, model, and
experiment (one directory per cell), so the per-cell tables in the main text
map directly onto the release; the experiments there carry short internal
identifiers (E1 = predicate-\test{} localization, E2 = router separability,
E3 = predicate generality, E6 = IOI contrast, E7 = the three-output rule,
E8 = multi-token answers, E9 = joint two-pair DAS).
The cross-model and cross-language analyses (internally E4/E5) are computed
from the stored E1 outputs and have no separate directories.

\paragraph{Compute and software.}
All experiments run on single workstation-class NVIDIA GPUs in
\texttt{bfloat16} via PyTorch and HuggingFace \texttt{transformers}; no run
requires more than one GPU. A localization cell
(one model $\times$ one language) takes roughly one to two GPU-hours on
Gemma-3-4B and proportionally more for the larger models; the router analysis
(greedy sweep + DAS training) takes one to two GPU-hours per model, and the
three-output run under one GPU-hour. No distributed training is involved;
model weights are never updated. Gemma-3 models are used under the Gemma
Terms of Use and Qwen3-8B under its Apache-2.0 license.

\section{Realistic Conditional Rules}
\label{app:realistic}

The studied template is the minimal shared skeleton of conditional
instructions that ship in real prompts. Each maps onto the same three roles:
a predicate over an input, a true-branch action, a false-branch action.

\begin{table}[!ht]
\centering
\small
\setlength{\tabcolsep}{3pt}
\begin{tabular}{@{}p{0.30\columnwidth}p{0.30\columnwidth}p{0.28\columnwidth}@{}}
\toprule
Deployed rule & Predicate $P(x)$ & Branches $A/B$ \\
\midrule
``If the request asks for medical advice, refuse; otherwise answer.'' &
topic of the request & \emph{refuse} / \emph{answer} \\[3pt]
``If the user mentions a refund, route to billing; else to support.'' &
intent of the message & \emph{billing} / \emph{support} \\[3pt]
``If the order total exceeds the limit, flag it; otherwise approve.'' &
numeric threshold & \emph{flag} / \emph{approve} \\
\bottomrule
\end{tabular}
\caption{Deployed prompt patterns instantiating ``if $P(x)$ then $A$ else
$B$.'' The third pattern is, up to wording, the studied rule, and is run
directly below.}
\label{tab:realistic}
\end{table}

Our canonical rule isolates this skeleton under controlled conditions:
single-token branch labels (relaxed to multi-token labels below) make the
flip metrics exact, and a numeric predicate makes truth assignment
unambiguous. What the findings imply for the deployed forms follows the two
stages. The \test{} band is separable, language-invariant, and generalizes
across predicate families, label pairs, a three-output rule, and multi-token
answers; this suggests the \emph{judgment} half of such rules is a stable
target for probing and portable steering. The \route{}
result cautions that any intervention built on the \emph{action} half is
token-bound: it will not survive a change of action vocabulary (\emph{refuse}
$\to$ \emph{decline}), though it does survive a change of prompt language.

\begin{figure}[htbp]
  \centering
  \includegraphics[width=\columnwidth]{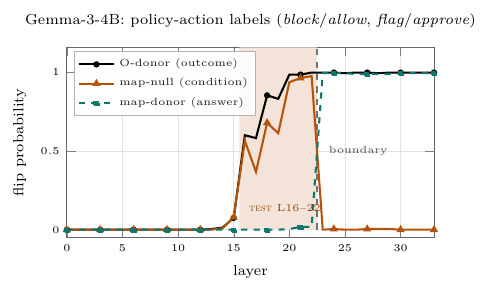}
  \caption{Policy-action labels (Gemma-3-4B, English;
  \emph{block}/\emph{allow} and \emph{flag}/\emph{approve} pooled): per-layer
  flip for the \odonor{} (outcome), \mapnull{} (condition) and \mapdonor{}
  (answer) donors. The shaded band (layers with \odonor{} flip $\ge0.6$ and
  \mapdonor{} flip $\le0.2$) contains the noun-label \test{} band, and the
  L22$\to$L23 stage boundary is identical.}
  \label{fig:policy}
\end{figure}

As a direct check, we reran the localization on the primary cell (English,
Gemma-3-4B) with two policy-\emph{action} pairs (\emph{block}/\emph{allow}
and \emph{flag}/\emph{approve}, both single tokens) in place of the noun
pairs, with no other change. The model solves the rule at accuracy $0.984$
($n{=}228$ gated items), and the localization reproduces exactly: \test{}
layer L20 with \odonor{} flip $0.99$, \mapdonor{} $0.00$, \mapnull{} $0.94$
(McNemar $p=3.3\times10^{-50}$), nulls $\le 0.031$, the same L22$\to$L23
stage boundary, and the same verdict as the grid cells
(\textsc{pass\_test\_isolated}; Figure~\ref{fig:policy}). The skeleton's
behavior is unchanged when its branches are the action words of the deployed
patterns above.

\begin{figure}[htbp]
  \centering
  \includegraphics[width=\columnwidth]{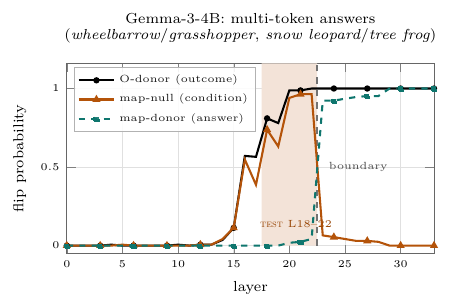}
  \caption{Multi-token answers (Gemma-3-4B, English; sub-word pair and
  two-word-phrase pair pooled, sequence-margin metrics): per-layer flip for
  the \odonor{} (outcome), \mapnull{} (condition) and \mapdonor{} (answer)
  donors. The shaded band (same criterion as Figure~\ref{fig:policy})
  contains the single-token \test{} band, with the same L22$\to$L23
  boundary.}
  \label{fig:multitok}
\end{figure}

A second check removes the single-token constraint itself. We generalized
the metrics to \emph{multi-token} answers (the flip becomes a
teacher-forced sequence-margin sign change, with the patch held at the
prompt's last-token position while candidate tokens are scored, and the mass
gate applied to the candidates' first tokens) and reran the localization
with two multi-token pair types pooled: sub-word-split single words
(\emph{wheelbarrow}/\emph{grasshopper}) and two-word phrases
(\emph{snow leopard}/\emph{tree frog}), each two tokens with distinct first
tokens. The model solves the rule at accuracy $0.964$ ($n{=}168$ gated), and
the localization reproduces once more: \test{} layer L20 (condition peak
L21), \odonor{} flip $0.99$, \mapdonor{} $0.02$, \mapnull{} $0.94$
(McNemar $p=6.8\times10^{-37}$), nulls $\le 0.04$, and the same
L22$\to$L23 boundary (Figure~\ref{fig:multitok}). Gemma-3's
$256$k vocabulary makes virtually every common English noun a single token,
so multi-token answers require sub-word-split or multi-word candidates by
construction.

Untested here, and required before these implications carry to production
rules: native-script actions, compositional predicates, and free-form rather
than constrained-choice completions
(\S\ref{sec:exp-generality}; Limitations).

\section{Recruit vs.\ Build: The IOI Contrast}
\label{app:ioi}

Is the conditional circuit \emph{recruited} from a pretrained mechanism or
\emph{built} at inference? We contrast it (English, Gemma-3-4B) with the
canonical weight-stored circuit, Indirect Object Identification (IOI)
\citep{ioi_circuit}: ``When John and Mary went to the store, John gave a drink
to'' $\to$ ``Mary''. With the same interchange machinery, we rank the IOI
\emph{mover} heads (those carrying the IO/S logit difference to the last
token under a role-swap corruption) and measure their overlap with our
conditional's heads.

\begin{figure}[htbp]
  \centering
  \includegraphics[width=\columnwidth]{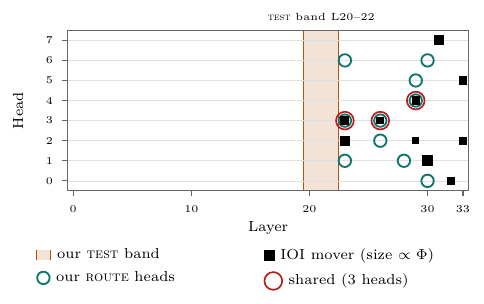}
  \caption{IOI-contrast head map (Gemma-3-4B): IOI mover heads (black,
  sized by $\Phi$) never fall in the \test{} band (amber); three coincide
  with our \route{} heads (teal circles; shared heads ringed red).}
  \label{fig:e6overlap}
\end{figure}

Two caveats frame one clean result (Figure~\ref{fig:e6overlap}).
First, the IOI movers are individually \emph{weak} here (the strongest
recovers only $\Phi=0.27$ of the swing, three of ten with zero OV
copy-score), so IOI in Gemma-3-4B is itself distributed and the mover ranking
is a weak-signal set; the contrast is suggestive, not decisive. Second, since
mover heads live late by construction, zero mid-stack overlap is partly
expected; the \emph{depth-matched} comparison is against the \route{} band,
and there the asymmetry appears: \emph{no} IOI mover falls in the
predicate-\test{} band L20--22 (overlap $0.00$), while the depth-matched
\route{} heads share $3$ of $10$ movers (L23H3, L26H3, L29H4; overlap $0.30$,
Jaccard $0.18$), consistent with the \route{} drawing on general late-layer
readout machinery that IOI also uses, rather than a dedicated recruited
circuit. Reported per head-set rather than collapsed to one number, the
verdict is \emph{\test: distinct (built) / \route: shared}: the same
\test{}/\route{} asymmetry the router analysis found
(\S\ref{sec:exp-router}), seen from a different angle.

\section{The Three-Output Rule}
\label{app:multilabel}

\begin{figure}[htbp]
  \centering
  \includegraphics[width=\columnwidth]{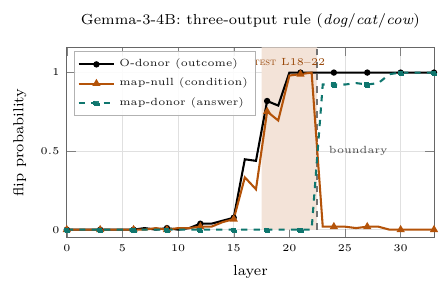}
  \caption{Three-output rule (Gemma-3-4B): per-layer flip for the \odonor{}
  (outcome), \mapnull{} (condition) and \mapdonor{} (answer) donors of the
  \emph{dog}/\emph{cat}/\emph{cow} variant. The shaded band (L18--22, same
  criterion as Figure~\ref{fig:policy}) contains the binary \test{} band, and
  the L22$\to$L23 stage boundary is identical: below it patches follow the
  condition (\odonor{}, \mapnull{}), above it the answer word (\mapdonor{}).}
  \label{fig:ml3}
\end{figure}

The three-output variant (\S\ref{sec:exp-multilabel}) uses three numeric
buckets with two thresholds ($x\ge5$, $2\le x\le4$, $x\le1$) mapped to
three single-token labels $A,B,C$ (\emph{dog}/\emph{cat}/\emph{cow}). In place
of the binary swap, the mapping is a cyclic rotation $r\in\{0,1,2\}$: bucket
$b$ takes label $(b{+}r)\bmod 3$, so the rule always contains the three label
words once each and every prompt tokenizes to the same length. Demonstrations
cover all three buckets. For a base query in the top bucket under the standard
mapping (answer $A$), with common flip target $B$, the four donors are:
\basecond{} ($r{=}0$, top bucket $\to A$); \odonor{} ($r{=}0$, middle bucket
$\to B$, a one-boundary move); \mapdonor{} ($r{=}1$, top bucket $\to B$,
rotated map, \emph{same} bucket); and \mapnull{} ($r{=}2$, middle bucket $\to
A$, rotated map whose answer returns to the base). A condition-carrying layer
follows the donor's bucket: the base's own rule then maps the middle bucket
to $B$, so \odonor{} and \mapnull{} flip while \mapdonor{} (same bucket) does
not; a readout layer follows the answer word, so \mapdonor{} and \odonor{}
flip while \mapnull{} does not. This is the binary isolation logic with a
rotation standing in for the swap. Figure~\ref{fig:ml3} shows the resulting
per-layer curves.

\section{Additional Figures}
\label{app:figs}

\begin{figure}[htbp]
  \centering
  \includegraphics[width=\columnwidth]{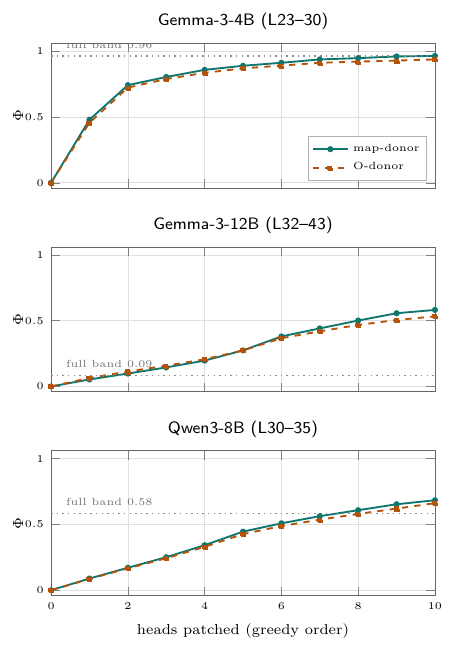}
  \caption{Greedy head-group patching: fraction-of-swing $\Phi$ as heads are
  added in greedy order, for the mapping and outcome donors (English). The two
  curves are nearly coincident in every model: the same heads, recruited in
  nearly the same order, carry both the mapping and the outcome, so no distinct
  routing heads emerge. Dotted line: patching the \emph{whole} band at once.
  Gemma-3-4B reaches $0.8\times$ the full-band effect with three heads;
  in Gemma-3-12B the greedy group far exceeds the full-band patch
  ($0.58$ vs.\ $0.09$), i.e.\ patching all band heads together largely cancels.}
  \label{fig:e2greedy}
\end{figure}

\begin{figure}[htbp]
  \centering
  \includegraphics[width=\columnwidth]{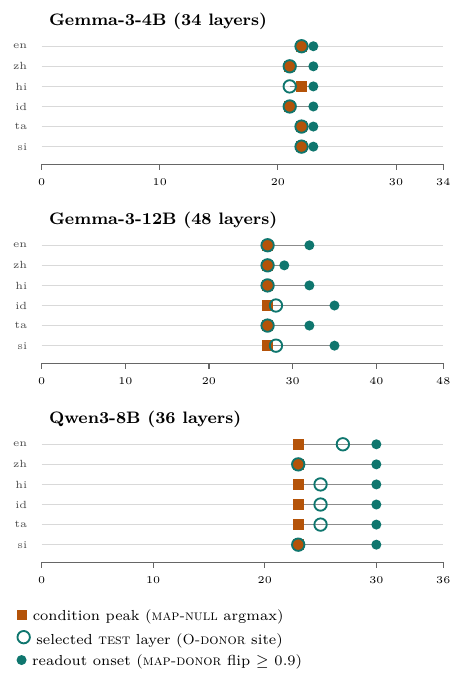}
  \caption{Condition peak (\mapnull{} argmax, amber square), selected
  \test{} layer (\odonor{} site, open teal circle), and readout onset
  (first \mapdonor{} flip $\ge0.9$, filled teal dot) for every cell,
  computed from the stored per-layer curves. Gemma-3-4B fuses all three;
  Qwen3-8B holds \emph{both} the condition (L23) and the readout onset (L30)
  language-invariant with the \test{} site between them; Gemma-3-12B anchors
  the condition (L27) but its readout onset varies by language (L29--L35).}
  \label{fig:bandpeaks}
\end{figure}

\begin{figure*}[htbp]
  \centering
  \includegraphics[width=\textwidth]{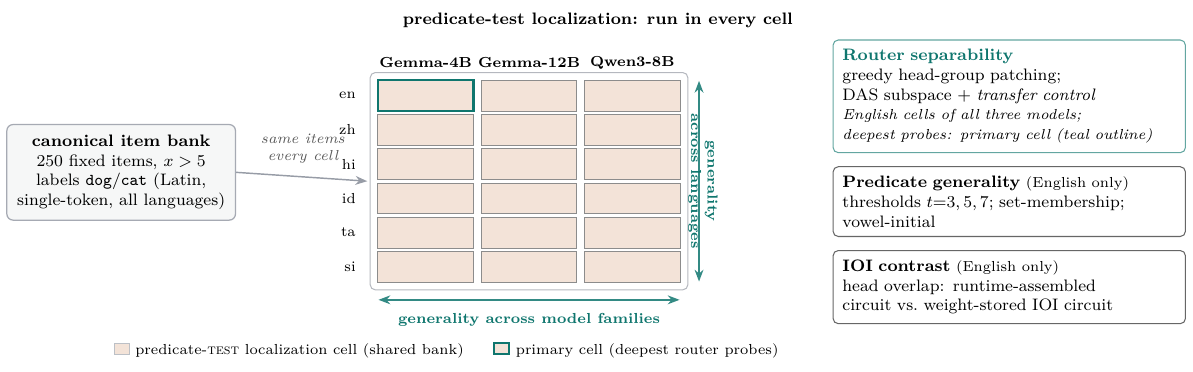}
  \caption{Experimental design. One fixed 250-item bank feeds every cell of the
  $6$-language $\times$ $3$-model localization grid (only the rule text is
  translated; the single-token Latin labels are held fixed). The router
  analysis runs on the English cells of all three models; the teal outline
  marks the primary cell (English, Gemma-3-4B), which also hosts the
  cross-lingual transfer probe and the label-generality replications.
  Predicate generality and the IOI contrast are English-only companion
  analyses.}
  \label{fig:design}
\end{figure*}

\begin{figure}[htbp]
  \centering
  \includegraphics[width=\columnwidth]{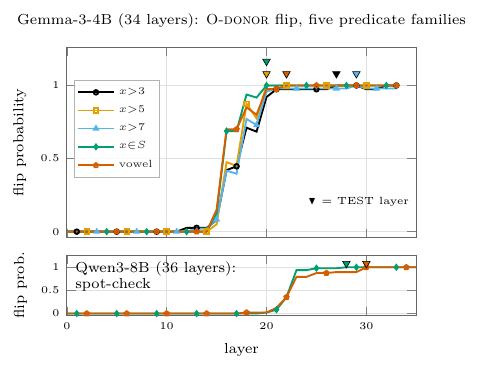}
  \caption{Predicate generality (Gemma-3-4B): per-layer \odonor{} flip for
  all five predicate families, each reaching $1.0$ in the mid-stack band;
  triangles mark selected \test{} layers. Bottom strip: Qwen3-8B spot-check
  (set-membership, vowel). Max null flip $0.062$.}
  \label{fig:e3gen}
\end{figure}

\begin{figure*}[htbp]
  \centering
  \includegraphics[width=\textwidth]{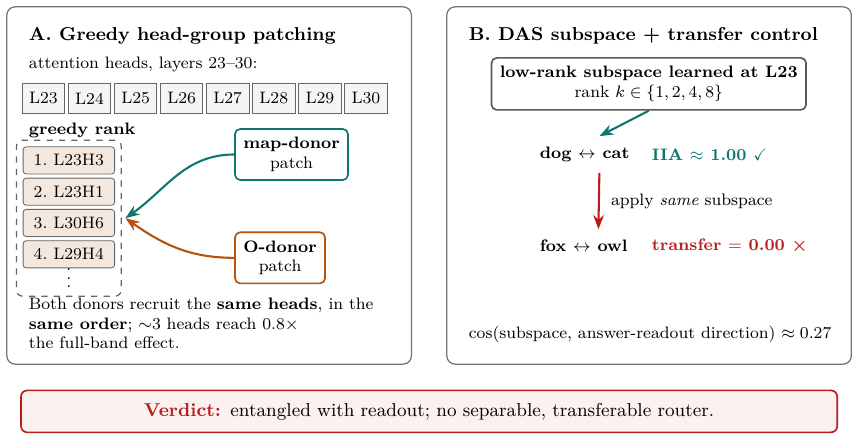}
  \caption{Two probes of the argument \route{} step, shown on the primary cell
  (Gemma-3-4B, English; see Table~\ref{tab:e2} for all three models).
  \emph{Left:} greedy head-group patching over L23--30 recruits the same heads
  in the same order for the mapping (\mapdonor{}) and the outcome (\odonor{})
  donors. \emph{Right:} a DAS-lite subspace at L23 flips dog\,$\leftrightarrow$\,cat
  ($\text{IIA}=1.00$) but does not transfer to a new label pair (fox/owl,
  $\approx0.01$; cosine with the answer-readout direction $0.27$--$0.39$).
  Verdict: \textsc{entangled\_with\_readout}.}
  \label{fig:e2}
\end{figure*}

Figures~\ref{fig:e2greedy}--\ref{fig:e2} provide the greedy head-accumulation
curves, the per-cell band geometry, the predicate-generality curves, the
experimental-design overview, and the router-methods schematic
referenced from the main text.

\begin{table}[htbp]
\centering
\small
\setlength{\tabcolsep}{2.5pt}
\begin{tabular}{@{}llccc@{}}
\toprule
Predicate & Domain & Behav.\ acc & \test{} layer & peak flip \\
\midrule
$x>3$          & numbers & 0.875 & L27 & 1.00 \\
$x>5$          & numbers & 1.000 & L20 & 1.00 \\
$x>7$          & numbers & 0.979 & L29 & 1.00 \\
$x\in\{2,3,5,7\}$ & numbers & 0.979 & L20 & 1.00 \\
vowel($x$)     & letters & 0.854 & L22 & 1.00 \\
\bottomrule
\end{tabular}
\caption{Predicate generality (Gemma-3-4B, English). Every predicate family
localizes a clean mid-stack \test{} band with peak outcome-flip $1.00$ and null
$\le 0.05$, including non-monotonic set-membership and the non-numeric vowel
predicate. A Qwen3-8B spot-check reproduces set-membership (L28) and vowel
(L30) at peak-flip $1.00$.}
\label{tab:e3}
\end{table}

\section{Per-Language Flip Curves}
\label{app:curves}

\begin{figure*}[htbp]
  \centering
  \includegraphics[width=\textwidth]{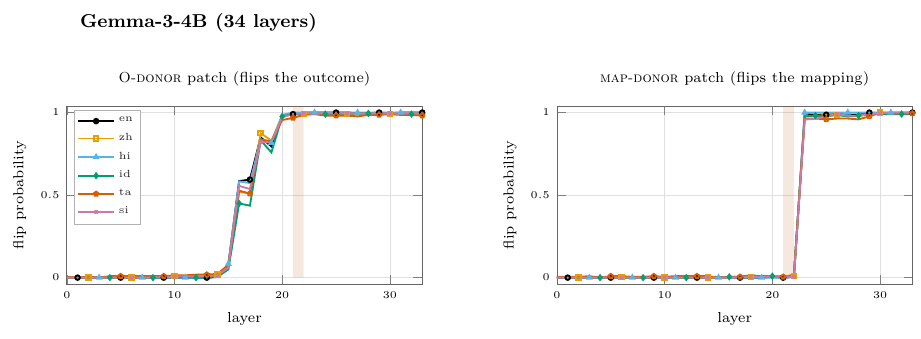}
  \caption{Gemma-3-4B: per-layer flip probability for all six languages.
  Left: \odonor{}; right: \mapdonor{}. Shading: range of selected \test{}
  layers (L21--22).}
  \label{fig:curves4b}
\end{figure*}

\begin{figure*}[htbp]
  \centering
  \includegraphics[width=\textwidth]{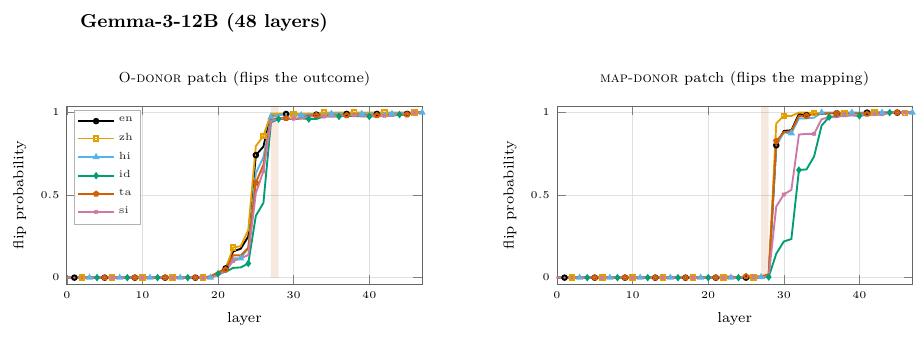}
  \caption{Gemma-3-12B: as above; \test{} band L27--28. The \mapdonor{} rise
  varies by language downstream of the boundary while the \odonor{} curves
  coincide.}
  \label{fig:curves12b}
\end{figure*}

\begin{figure*}[htbp]
  \centering
  \includegraphics[width=\textwidth]{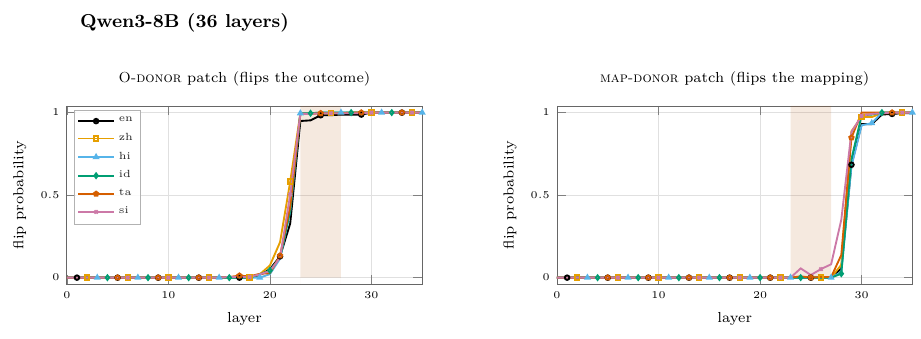}
  \caption{Qwen3-8B: as above; \test{} band L23--27. The \mapdonor{} rise is
  displaced several layers past the band: the staggered
  condition$\to$readout structure.}
  \label{fig:curvesqwen}
\end{figure*}

Figures~\ref{fig:curves4b}--\ref{fig:curvesqwen} plot the per-layer flip
probabilities behind Table~\ref{tab:e1}: one panel pair per model, six
languages overlaid, with the \odonor{} patch left, the \mapdonor{} patch
right, and shading marking the selected \test{} layers. They show the
per-language coincidence of the \test{} band, the stage boundary, and the
readout-onset geometry of \S\ref{sec:exp-generality}.

\end{document}